\documentclass[10pt,twocolumn,letterpaper]{article}

\usepackage{cvpr}      
\usepackage[accsupp]{axessibility}
\definecolor{cvprblue}{rgb}{0.21,0.49,0.74}
\usepackage[pagebackref,breaklinks,colorlinks,allcolors=cvprblue]{hyperref}

\def\paperID{43} 
\def\confName{CVPR}
\def\confYear{2026}

\title{T-Gated Adapter: A Lightweight Temporal Adapter for Vision-Language Medical Segmentation}

\author{Pranjal Khadka\\
Independent Researcher\\
{\tt\small pranzalkhadka1@gmail.com}
}

\begin{document}
\maketitle
\begin{abstract}
Medical image segmentation traditionally relies on fully supervised 3D architectures that demand a large amount of dense, voxel-level annotations from clinical experts which is a prohibitively expensive process. Vision Language Models (VLMs) offer a powerful alternative by leveraging broad visual semantic representations learned from billions of images. However, when applied independently to 2D slices of a 3D scan, these models often produce noisy and anatomically implausible segmentations that violate the inherent continuity of anatomical structures. We propose a temporal adapter that addresses this by injecting adjacent-slice context directly into the model’s visual token representations. The adapter comprises a temporal transformer attending across a fixed context window at the token level, a spatial context block refining within-slice representations, and an adaptive gate balancing temporal and single-slice features. Training on 30 labeled volumes from the FLARE22  dataset, our method achieves a mean Dice of 0.704 across 13 abdominal organs with a gain of +0.206 over the baseline VLM trained with no temporal context. Zero-shot evaluation on BTCV and AMOS22 datasets yields consistent improvements of +0.210 and +0.230, with the average cross-domain performance drop reducing from 38.0\% to 24.9\%. Furthermore, in a cross-modality evaluation on AMOS22 MRI with neither model receiving any MRI supervision, our method achieves a mean Dice of 0.366, outperforming a fully supervised 3D baseline (DynUNet, 0.224) trained exclusively on CT, suggesting that CLIP's visual semantic representations generalize more gracefully across imaging modalities than convolutional features.
\end{abstract}    
\section{Introduction}
\label{sec:intro}

Recent advancements in vision–language segmentation models such as CLIPSeg~\cite{luddecke2022image}, built on top of CLIP~\cite{radford2021learning} have introduced a new paradigm for image segmentation. Unlike traditional fully supervised models such as U-Net~\cite{ronneberger2015u} and nnU-Net~\cite{isensee2021nnu} that require training on specific, pre-defined class labels, VLMs offer the capability of zero-shot or few-shot segmentation via text prompts. This flexibility is particularly promising for medical imaging, where acquiring large-scale, pixel-level annotations for every organ and pathology is prohibitively expensive~\cite{tajbakhsh2020embracing}.

A defining hallmark of VLMs is their capacity for generalization beyond their training distribution. In the natural image domain, these models routinely demonstrate zero-shot transfer to unseen object categories by relying on robust semantic representations rather than memorized pixel statistics~\cite{jia2021scaling}. In medical imaging, this semantic grounding can address one of the field's most persistent challenges: cross-modality domain shift. Traditional 3D convolutional networks overfit to the physical intensity characteristics of their training modality and fail when those statistics change~\cite{guan2021domain}. VLMs, by contrast, encode high level anatomical concepts that are not inherently tied to any particular imaging physics. Therefore, a VLM trained on one modality of data should be much more capable of generalizing to a different modality, navigating a fundamental shift in image formation that typically renders task-specific models unusable.

\begin{figure*}[t]
    \centering
    \includegraphics[width=\textwidth]{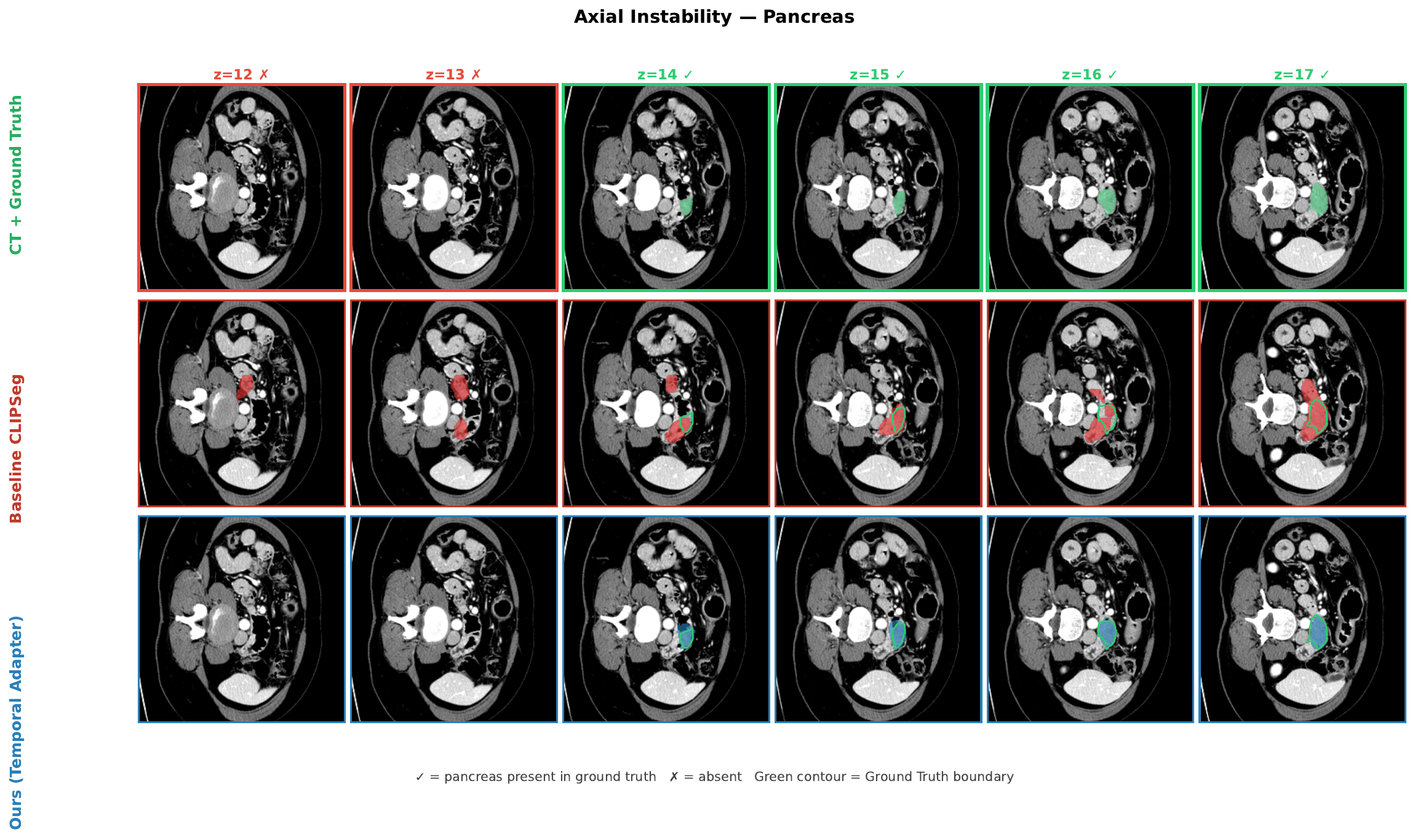}
    \caption{Temporal inconsistency of slice-wise VLM inference on the pancreas. Green slice labels indicate the organ is present in ground truth. The baseline (middle row) produces spurious detections in anatomically implausible slices. Our method (bottom row) remains consistent, suppressing predictions where the pancreas is absent.}
    \label{fig:temporal}
\end{figure*}

However, a fundamental domain gap exists between standard VLMs and clinical workflow: current foundation models are predominantly 2D-native, whereas medical imaging modalities such as Computed Tomography (CT) and Magnetic Resonance Imaging (MRI) are inherently three-dimensional volumes. As a result, a common strategy is to decompose 3D volumes into sequences of 2D axial slices, which are then processed independently by 2D vision or vision–language models~\cite{ma2024segment}. This slice-wise adaptation enables the reuse of powerful 2D foundation models, but comes at the cost of discarding volumetric context along the axial direction. By treating adjacent slices as independent and identically distributed (i.i.d.) samples, such approaches lose the ability to enforce anatomical continuity. In segmentation tasks, this often manifests as spatial
instability where structures appear inconsistently across neighboring slices~\cite{wang2025sam} as illustrated in Figure~\ref{fig:temporal}.

We propose a temporal adapter that injects adjacent-slice context into the model’s visual token representations during training. Given a center slice and its four nearest neighbors, all five slices are encoded by the CLIP vision encoder and a temporal transformer attends across the slice dimension independently for each token position. Each spatial location in the center slice can aggregate evidence from the corresponding location in neighboring slices, learning to suppress detections that lack cross-slice anatomical support. A spatial self-attention block then refines these features within the slice, and an adaptive gate blends the volumetrically-informed features with the original single-slice representations, allowing conservative use of temporal context where it is most beneficial.

We evaluate on FLARE22~\cite{ma2024unleashing}, training on 30 labeled volumes and testing on a held-out set of 10. To confirm improvements are not specific to FLARE22, we additionally evaluate zero-shot on BTCV~\cite{landman2015miccai} and AMOS22~\cite{ji2022amos} with no model adaptation of any kind.

Our contributions are as follows:

\begin{itemize}
\item We propose a temporal adapter that injects adjacent-slice context into 2D vision-language models at the token level, enabling 3D-aware segmentation without modifying the base model.

\item We demonstrate that the proposed adapter substantially reduces cross-domain performance drop from 38.0\% to 24.9\% under zero-shot transfer to two independent CT benchmarks, indicating that the model learns genuine volumetric understanding rather than dataset-specific patterns.

\item In a zero-shot cross-modality evaluation on AMOS22 MRI, our approach achieves a mean Dice of 0.366, outperforming a supervised 3D baseline (DynUNet, 0.224) that also receives no MRI supervision, suggesting that CLIP-based representations generalize more robustly across imaging modalities than task-specific convolutional features.
\end{itemize}

\section{Related Work}
\label{sec:related}

\paragraph{Medical Image Segmentation.}
Convolutional encoder-decoder architectures have long dominated medical image segmentation, beginning with the foundational UNet. To leverage the rich volumetric context of clinical scans, three-dimensional extensions such as 3D U-Net~\cite{cciccek20163d} process full CT and MRI volumes natively using 3D convolutions. More recently, transformer-based architectures such as UNETR~\cite{hatamizadeh2022unetr} and Swin UNETR~\cite{hatamizadeh2021swin} have established new state-of-the-art benchmarks by capturing long-range spatial dependencies within volumetric data. However, while these specialized architectures achieve state-of-the-art precision on in-domain data, they suffer from two practical limitations: they require large datasets of dense 3D annotations to avoid overfitting, and their closed-set formulation restricts them to a fixed taxonomy of organs defined prior to training. Our work seeks to achieve competitive volumetric segmentation without the high data constraints of training 3D architectures from scratch.

\paragraph{Foundation Models in Medical Vision.}
The introduction of large-scale visual foundation models, particularly VLMs like CLIP and promptable architectures like SAM~\cite{kirillov2023segment}, has shifted the paradigm toward zero-shot and few-shot inference. In the medical domain, models such as MedSAM~\cite{ma2024segment} adapt geometric prompting to clinical imaging, while CLIPSeg utilizes text embeddings to segment objects via natural language queries. The landscape of medical VLMs has also rapidly expanded with the introduction of domain-specific contrastive models like BiomedCLIP~\cite{zhang2023large} and RadCLIP~\cite{lu2025radclip}, which align medical images with radiological reports. While these models offer unprecedented semantic robustness and data-efficiency, they are inherently designed for 2D images. Consequently, applying them to 3D medical volumes on a slice-by-slice basis discards crucial anatomical continuity, leading to spatial instability~\cite{gong20243dsam}.

\paragraph{Data-Efficient 3D Adaptation.}
To extend 2D models to 3D domains without the high data requirements of training volumetric architectures from scratch, recent works have increasingly leveraged foundation models~\cite{wu2025medical}. In medical imaging, extending 2D foundation models to 3D has largely focused on recurrent connections or heavy volumetric upsampling~\cite{chen2016combining}. In contrast, our approach introduces a lightweight, factorized temporal-spatial adapter equipped with an adaptive gate. This allows a fine-tuned 2D VLM to aggregate adjacent-slice context efficiently, maintaining its semantic priors while dynamically resolving the spatial ambiguities of single slice inference.
\section{Method}
\label{sec:method}

\subsection{Base Model}
\begin{figure*}[t]
    \centering
    \includegraphics[width=\textwidth]{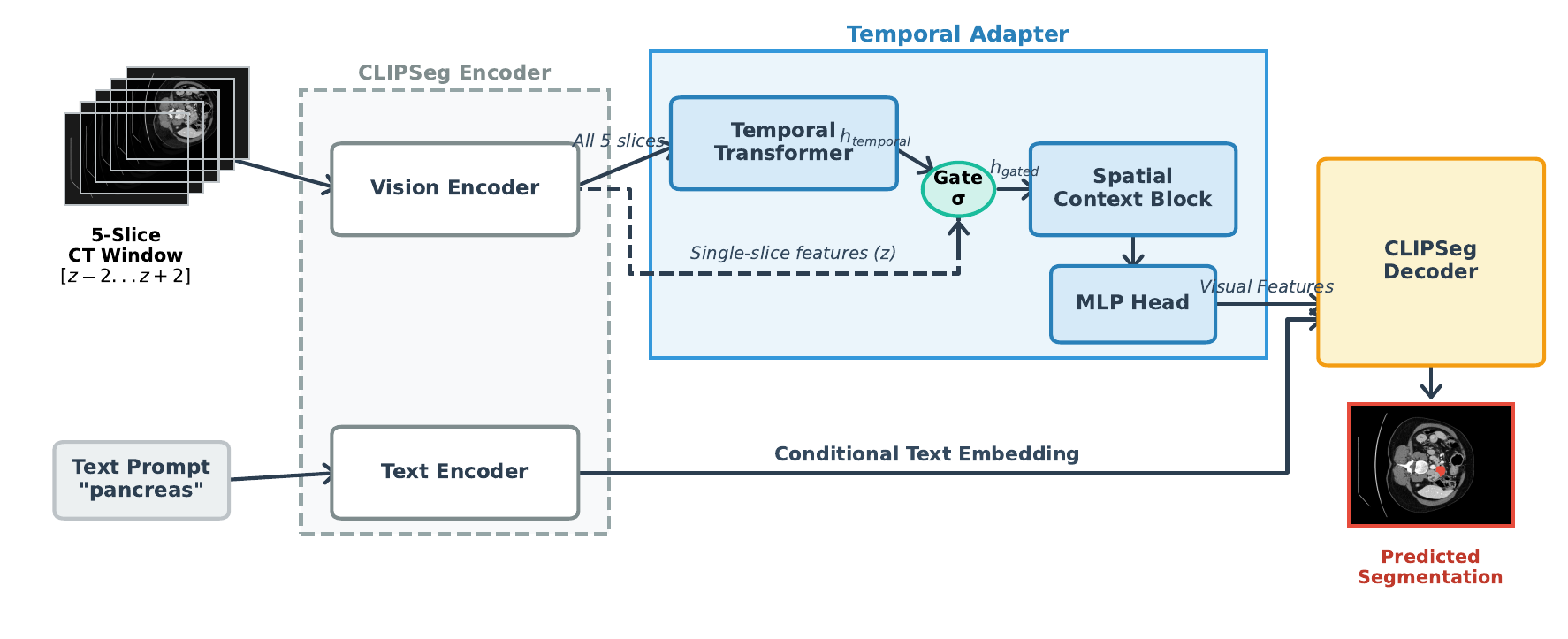}
    \caption{Overview of the proposed temporal adapter. Novel components include temporal transformer, spatial context block, and adaptive gate which are highlighted in blue and green within the \textit{Temporal Adapter} module.}
    \label{fig:architecture}
\end{figure*}

Figure~\ref{fig:architecture} provides an overview of our proposed architecture. CLIPSeg consists of a CLIP ViT-B/16 vision encoder, a CLIP text encoder, and a lightweight transformer decoder. Given a 2D image and a text prompt, the vision encoder produces $L$ patch tokens of dimension $D_v$, the text encoder produces a conditional embedding, and the decoder generates a binary segmentation map. Applied slice-by-slice to CT volumes, this model produces temporally inconsistent predictions due to the complete absence of inter-slice information in the visual token representations. We fine-tune CLIPSeg on abdominal CT segmentation using binary cross-entropy and Dice loss with equal weighting, with differential learning rates: $10^{-6}$ for vision and text encoders, $10^{-5}$ for the decoder, and $5\times10^{-5}$ for all adapter parameters.

\subsection{Temporal Transformer}
Given a center slice $z$ and a 5-slice context window $\mathcal{W} = [z{-}2,\, z{-}1,\, z,\, z{+}1,\, z{+}2]$, the CLIP vision encoder processes all five slices in parallel, producing token sequences of shape $(5, L, D_v)$. We reshape this to $(L, 5, D_v)$, treating each spatial token position independently across the slice dimension. A learned linear projection reduces the token dimension to $D_\text{proj} = 256$, followed by layer normalization and learned temporal position embeddings encoding slice order within the window:
\begin{equation}
    \mathbf{x}_\text{proj} = \text{LN}\!\left(\mathbf{W}_\text{in}\,\mathbf{x}\right) + \mathbf{e}_\text{pos},
    \label{eq:proj}
\end{equation}
where $\mathbf{e}_\text{pos} \in \mathbb{R}^{1 \times 5 \times D_\text{proj}}$ is a learned parameter.

A stack of $N{=}4$ transformer encoder layers with pre-norm architecture and stochastic depth regularization (drop-path rate increasing linearly from $0$ to $0.1$ across layers) attends across the 5-slice dimension for each token position. Each spatial location aggregates evidence from the corresponding location in neighboring slices, learning to suppress activations that lack cross-slice anatomical support. The output is projected back to $D_v$ and center-slice features are extracted, enriched with volumetric context.

\subsection{Spatial Context Block}
The temporal transformer operates independently at each spatial token position and does not model within-slice spatial relationships. A subsequent spatial self-attention block addresses this by attending across all $L$ token positions of the center slice, allowing spatially adjacent tokens to share the volumetric information gathered by the temporal transformer. This produces a globally coherent representation before the decoder, using the same pre-norm architecture with a two-layer MLP feedforward network and GELU activations.

\subsection{Adaptive Gate Mechanism}
The benefit of temporal context varies across structures and slices. For large organs such as the liver, single-slice features are sufficient and aggressive temporal fusion may introduce noise. We introduce a learned gate that interpolates between temporally fused features $\mathbf{h}_\text{temporal}$ and the original single-slice features $\mathbf{h}_\text{single}$:
\begin{equation}
    \mathbf{h}_\text{center} = g \cdot \mathbf{h}_\text{temporal} + (1 - g) \cdot \mathbf{h}_\text{single},
    \label{eq:gate}
\end{equation}
where $g = \sigma(\mathbf{W}_g \mathbf{h}_\text{temporal} + \mathbf{b}_g)$. The weight $\mathbf{W}_g$ is initialized to zero and bias $\mathbf{b}_g$ to $-5.0$, so the gate starts near zero and the model initially behaves identically to the CLIPSeg baseline. To prevent the model from defaulting to a simple averaging of features, we introduce a binary gating penalty $\lambda(g \odot (1 - g))$ with $\lambda = 0.001$, which encourages the gate to make decisive choices regarding the utility of temporal context.

\subsection{Training Details}
\label{sec:training}
 
\noindent\textbf{Data preprocessing.}
CT volumes are preprocessed with HU windowing $[-125, 275]$, normalized to $[0, 255]$, and resampled to $352{\times}352$ pixels per slice. Each training sample consists of a center slice paired with its four nearest axial neighbors as a 5-slice context stack; at volume boundaries, edge slices are replicated to maintain a fixed context size of 5.
 
\noindent\textbf{Negative sampling.}
Slices in which the queried organ is entirely absent are included as negative training samples with a fixed ratio of 1 negative per 3 positive slices. Negative samples are critical for learning temporal consistency: they are precisely the slices where slice-wise models hallucinate false positives, and exposing the model to these cases during training directly supervises suppression of cross-slice noise.
 
\noindent\textbf{Class-imbalanced sampling.}
Standard uniform sampling caused the model to neglect small and rare organs during training. Hence, weighted sampling was applied by assigning higher sampling probabilities to underrepresented structures.
Table~\ref{tab:sampling} reports the sampling weights for organs.
\begin{table}[h]
    \centering
    \small
    \caption{Class-imbalanced sampling weights}
    \label{tab:sampling}
    \begin{tabular}{lc}
        \toprule
        \textbf{Organ} & \textbf{Weight} \\
        \midrule
        R.\ Adrenal, L.\ Adrenal & $\times 8.0$  \\
        Duodenum, Esophagus      & $\times 8.0$  \\
        Pancreas, Stomach        & $\times 2.0$  \\
        Gallbladder              & $\times 2.0$  \\
        Liver, Spleen, Kidneys   & $\times 1.0$  \\
        Aorta, IVC               & $\times 1.0$  \\
        \bottomrule
    \end{tabular}
\end{table}

\noindent\textbf{Augmentation.}
Random rotation of $\pm5^\circ$ is applied to all organs. Random horizontal flipping is applied only to non-lateralized organs, preserving left-right anatomical identity for kidneys and adrenal glands. The same augmentation is applied consistently across all 5 slices in a context stack to maintain geometric coherence.
 
\noindent\textbf{Optimization.}
Training uses AdamW~\cite{loshchilov2017decoupled} with weight decay $10^{-4}$ and differential learning rates: $10^{-6}$ for vision and text encoders, $10^{-5}$ for the CLIPSeg decoder, and $5{\times}10^{-5}$ for all adapter parameters. The scheduler uses cosine annealing warm restarts with $T_0 = 5$ epochs. We train for 30 epochs with a batch size of 8, shuffling slices across all volumes each epoch. The checkpoint with the highest mean validation Dice is selected for evaluation. All training uses full float32 precision on a single NVIDIA T4 GPU.
\section{Experiments}
\label{sec:experiments}

\subsection{Experimental Setup}
FLARE22 provides abdominal CT volumes with 13 organ annotations: liver, right and left kidney, spleen, pancreas, aorta, inferior vena cava, right and left adrenal gland, gallbladder, esophagus, stomach, and duodenum. We use 30 volumes for training, 10 for validation, and 10 for held-out testing. BTCV and AMOS22 CT serve as zero-shot cross-domain CT benchmarks: 10 randomly sampled volumes from
each are evaluated with no model adaptation of any kind. For the cross-modality experiment, we additionally evaluate on 10 randomly sampled volumes from the AMOS22 MRI subset, applying both our method and a fully supervised 3D baseline DynUNet with no MRI supervision. Per-organ volumetric Dice is computed on each test volume.

\subsection{Comparison to Baseline}

Table~\ref{tab:main} reports mean Dice across all evaluation datasets. The fine-tuned CLIPSeg baseline with identical training setup but without the temporal adapter achieves $0.497$ on FLARE22. Adding the temporal adapter improves this to $0.704$, a gain of $+0.206$ and a $41\%$ relative improvement. The improvement is consistent across both zero-shot cross-domain CT benchmarks: $+0.210$ on BTCV and $+0.230$ on AMOS22. The average cross-domain drop decreases from $38.0\%$ to $24.9\%$, indicating that the adapter improves genuine volumetric understanding rather than in-domain fitting. Both models are trained on identical data with identical supervision --- the temporal adapter is the only difference.

\begin{table}[h]
    \centering
    \caption{Mean Dice across 13 abdominal organs on CT benchmarks. BTCV 
    and AMOS22 CT are zero-shot with no model adaptation. Average drop 
    computed relative to FLARE22. Both models trained on 30 labeled volumes.}
    \label{tab:main}
    \resizebox{\columnwidth}{!}{%
    \begin{tabular}{lcccc}
        \toprule
        \textbf{Method} & \textbf{FLARE22} & \textbf{BTCV} & \textbf{AMOS22 CT} \\
        \midrule
        CLIPSeg Baseline          & 0.497 & 0.334 & 0.283 \\
        CLIPSeg + Temporal (Ours) & \textbf{0.704} & \textbf{0.544} & \textbf{0.513} \\
        \bottomrule
    \end{tabular}}
\end{table}

\begin{figure*}[t]
    \centering
    \includegraphics[width=0.95\textwidth]{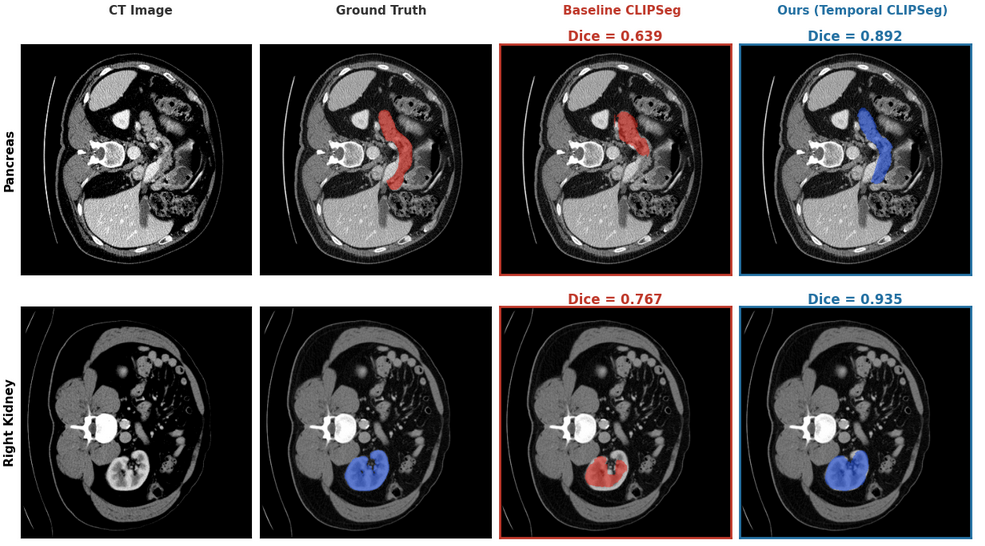}
    \caption{Qualitative segmentation comparison on the best-performing slice per organ. Dice scores are shown above each prediction column.}
    \label{fig:qualitative}
\end{figure*}

\subsection{Per-Organ Analysis}

Table~\ref{tab:perorgan} shows per-organ Dice on FLARE22. The largest
improvements occur on structures known to exhibit the most severe temporal inconsistency under slice-wise inference: pancreas ($+0.404$), stomach ($+0.268$), gallbladder ($+0.273$), and right kidney ($+0.337$). The pancreas in particular spans few axial slices and has high shape variability across patients; without cross-slice context, the model frequently produces false positives in adjacent slices where the pancreas is absent. Large visually distinctive organs such as the liver ($+0.049$) and aorta ($+0.171$) see smaller but consistent gains as their single-slice detectability was already strong, so the marginal contribution of volumetric context is smaller. The esophagus shows a regression ($-0.143$), consistent with its thin tubular morphology: the esophagus is nearly absent in many axial slices within any 5-slice window, and cross-slice attention imports noise from slices where it is not visible. Qualitative comparisons in Figure~\ref{fig:qualitative} demonstrate that our temporal adapter resolves these spatial ambiguities, producing contiguous and anatomically accurate boundaries compared to the baseline.

\begin{table}[h]
    \centering
    \caption{Per-organ Dice on FLARE22 test set (10 volumes). Mean is computed as volume-weighted average across all organs and volumes.}
    \label{tab:perorgan}
    \setlength{\tabcolsep}{8pt}
    \begin{tabular}{lccc}
        \toprule
        \textbf{Organ} & \textbf{Baseline} & \textbf{+ Temporal} & \textbf{$\Delta$ Dice} \\
        \midrule
        Liver          & 0.911 & 0.960 & $+0.049$ \\
        Spleen         & 0.691 & 0.919 & $+0.228$ \\
        Pancreas       & 0.243 & 0.647 & $+0.404$ \\
        Stomach        & 0.581 & 0.849 & $+0.268$ \\
        Aorta          & 0.706 & 0.877 & $+0.171$ \\
        Gallbladder    & 0.442 & 0.715 & $+0.273$ \\
        Esophagus      & 0.524 & 0.381 & $-0.143$ \\
        Duodenum       & 0.304 & 0.494 & $+0.190$ \\
        R.\ Adrenal    & 0.238 & 0.380 & $+0.142$ \\
        L.\ Adrenal    & 0.311 & 0.396 & $+0.085$ \\
        IVC            & 0.572 & 0.767 & $+0.195$ \\
        R.\ Kidney     & 0.499 & 0.836 & $+0.337$ \\
        L.\ Kidney     & 0.731 & 0.925 & $+0.194$ \\
        \midrule
        \textbf{Mean}  & 0.497 & \textbf{0.704} & $+0.206$ \\
        \bottomrule
    \end{tabular}
\end{table}

\subsection{Ablation: Text Prompt Sensitivity}
\label{sec:ablation}

A potential concern with VLM-based segmentation is that the model may
learn to ignore the text prompt and instead act as a generic visual
segmentor conditioned on spatial priors (\eg, always predicting a
blob in the upper-right quadrant for any query). To test whether the
model is genuinely conditioned on language, we evaluate two prompt
corruption conditions on the FLARE22 test set.

\noindent\textbf{Blank prompt.} The organ name is replaced with an
empty string (\texttt{""}), removing all semantic content from the
text input. As shown in Table~\ref{tab:ablation}, mean Dice collapses from $0.704$ to $0.005$, with 10 of 13 organs scoring exactly $0.000$.

\noindent\textbf{Wrong prompt.} Each organ is queried with a
semantically unrelated organ name (\eg, the liver is queried as
\texttt{"aorta"}, the pancreas as \texttt{"liver"}). Mean Dice
collapses to $0.011$, with 11 of 13 organs scoring below $0.001$.

\begin{table}[h]
    \centering
    \caption{Text prompt sensitivity on FLARE22 test set. Corrupting
    the text prompt causes near-total collapse, confirming that the
    model is conditioned on the language query.}
    \label{tab:ablation}
    \resizebox{\columnwidth}{!}{%
    \begin{tabular}{lcc}
        \toprule
        \textbf{Condition} & \textbf{Prompt Example} & \textbf{Mean Dice} \\
        \midrule
        Correct prompt & \texttt{"liver"} $\to$ liver GT   & $0.704$ \\
        Blank prompt   & \texttt{""} $\to$ liver GT         & $0.005$ ($-99.3\%$) \\
        Wrong prompt   & \texttt{"aorta"} $\to$ liver GT    & $0.011$ ($-98.4\%$) \\
        \bottomrule
    \end{tabular}}
\end{table}


\subsection{Cross-Modality Generalization}
 
We also conduct a zero-shot evaluation on the AMOS22 MRI subset which is a fundamentally different imaging modality with distinct tissue contrast. No model is exposed to MRI data at any point during fine-tuning. MRI volumes are preprocessed using per-volume percentile normalization (1st--99th percentile clipped to $[0, 255]$) to produce images comparable in dynamic range to CT slices.
 
\begin{figure*}[t]
    \centering
    \includegraphics[width=0.95\textwidth]{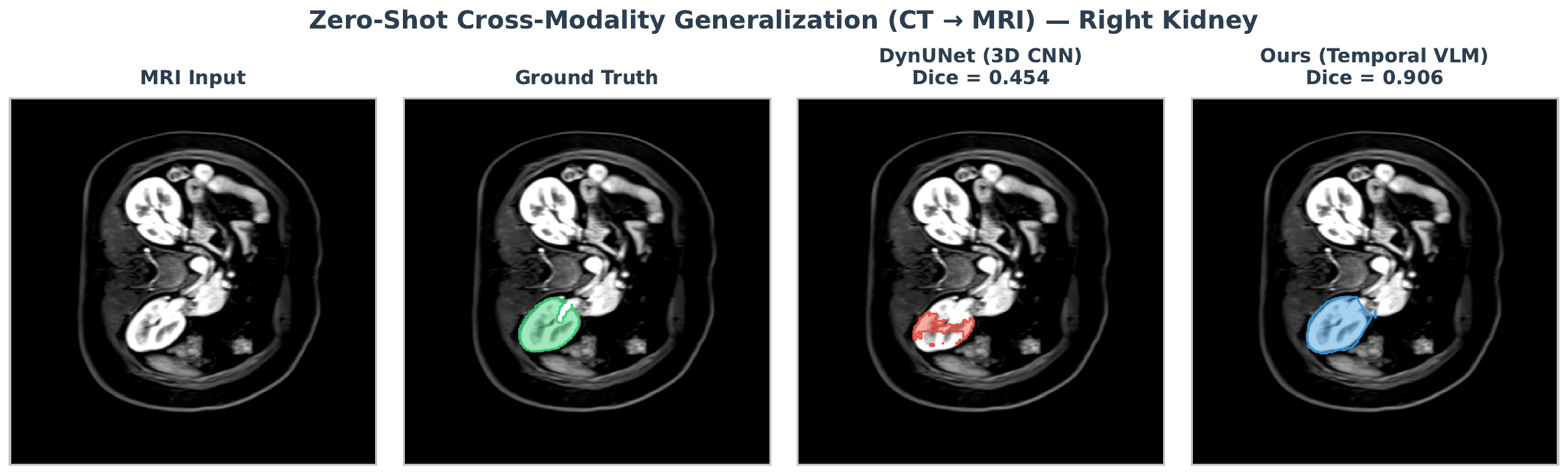}
    \caption{Zero-shot cross-modality generalization on AMOS22 MRI 
    (right kidney) with neither model receiving any MRI supervision.}
    \label{fig:cross_modality}
\end{figure*}

As shown in Table~\ref{tab:mri}, our method achieves a mean Dice of 0.366 on AMOS22 MRI, outperforming a supervised 3D baseline (DynUNet, 0.224) trained on the identical 30 FLARE22 CT volumes. The sharp decline of the 3D CNN highlights a known limitation: standard convolutional networks tightly fit the intensity distributions of their training modality, making them highly sensitive to the shift from CT to MRI. Our approach is less affected by this shift, suggesting that the foundational representations inherited from CLIP provide a degree of modality invariance as shown in Figure~\ref{fig:cross_modality}. By relying on broader semantic features rather than exact pixel values, the model retains better anatomical recognition even when the underlying imaging physics change entirely.

\begin{table}[h]
    \centering
    \caption{Zero-shot cross-modality evaluation on AMOS22 MRI (10 volumes) with neither model receiving any MRI supervision}
    \label{tab:mri}
    \resizebox{\columnwidth}{!}{%
    \begin{tabular}{lcc}
        \toprule
        \textbf{Method} & \textbf{AMOS22 MRI} \\
        \midrule
        DynUNet      & 0.224 \\
        CLIPSeg + Temporal Adapter    & \textbf{0.366} \\
        \bottomrule
    \end{tabular}}
\end{table}


\section{Discussion and Limitation}
\label{sec:discussion}

The per-organ improvement pattern provides clear evidence that our temporal adapter effectively resolves the spatial instability of 2D slice-wise inference. Structures with high morphological variance across slices, such as the pancreas and stomach, saw the most significant gains because the adapter enables the model to verify anatomical boundaries using adjacent-slice context. Conversely, the minor performance regression on the esophagus highlights a structural trade-off: for very thin, tubular organs, the target structure frequently disappears from the local context window, causing temporal attention to introduce background noise rather than useful signal. Overall, the fact that these performance trends remained stable during zero-shot evaluation on BTCV and AMOS22 confirms that the injected volumetric context is anatomically grounded, allowing the model to generalize more robustly beyond the training distribution than the baseline.

The cross-modality result on AMOS22 MRI further reinforces this conclusion. The fact that our model outperforms a supervised 3D baseline on MRI without any modality specific adaptation is consistent with the qualitative difference between how the two model families represent anatomy. DynUNet's learned features are tightly coupled to CT intensity distributions; when those statistics change, the representations lose their discriminative power. Our model inherits CLIP's language-grounded visual features, which encode semantic concepts robust enough to remain partially discriminative under the CT-to-MRI modality shift.

\paragraph{Limitations.}
Several limitations bound the scope of these results. First, the temporal adapter uses a fixed 5-slice context window regardless of slice spacing, which varies substantially across CT acquisitions. A dynamic window size adapted to the physical spacing of each volume would be a more principled design. Second, CLIPSeg requires resizing native CT slices from 512×512 to 352×352 pixels, and this downsampling discards fine spatial detail before the model processes the image. For small structures such as the adrenal glands, whose signal at native resolution is already limited, this resolution bottleneck constrains performance in a way the temporal adapter cannot compensate for.

\paragraph{Future Directions.}
Two directions follow naturally from the current limitations. First, 
a metadata-aware context window that adapts its temporal depth to the 
physical $z$-spacing extracted from DICOM headers would be more 
principled than the fixed 5-slice design used here. Second, applying 
this spatial-temporal adapter to higher-resolution 2D foundation models such as SAM would bypass the resolution bottleneck that currently limits performance on small structures.
\section{Conclusion}
\label{sec:conclusion}

Vision-language models offer a data-efficient paradigm for medical image segmentation by leveraging rich semantic priors instead of relying solely on massive annotated datasets. However, directly applying these 2D foundation models to volumetric scans slice-by-slice intrinsically discards essential anatomical continuity, leading to fragmented and unreliable predictions. To bridge this gap, we introduced a temporally-gated adapter that seamlessly injects adjacent-slice context into the visual representations of the model. By aggregating cross-slice evidence and refining it spatially, our lightweight module mitigates the spatial instability of 2D inference. Trained on only 30 labeled CT volumes, our approach achieved a substantial +0.206 mean Dice improvement over the baseline VLM on FLARE22. Furthermore, these gains remained consistent under zero-shot cross-domain evaluation on BTCV and AMOS22, confirming that the adapter learns genuine volumetric structure rather than overfitting to specific textures. In a cross-modality evaluation, our approach ($0.366$) outperforms a supervised 3D baseline that also receives no MRI training ($0.224$), suggesting that language-grounded representations generalize more robustly across imaging modalities than convolutional features learned from modality-specific intensity distributions. Overall, this work presents a data-efficient approach to leverage powerful 2D foundation models for the demanding 3D requirements of clinical imaging.

\medskip\noindent Code is available at \url{https://github.com/pranzalkhadka/T-Gated-Adapter}
{
    \small
    \bibliographystyle{ieeenat_fullname}
    \bibliography{main}
}


\end{document}